\documentclass[letterpaper]{article} 
\usepackage{aaai24}  
\usepackage{times}  
\usepackage{helvet}  
\usepackage{courier}  
\usepackage[hyphens]{url}  
\usepackage{graphicx} 
\usepackage{natbib}  
\usepackage{caption} 
\usepackage{algorithm}
\usepackage{algorithmic}

\usepackage{amsmath}
\usepackage{multirow}
\usepackage{makecell}

\DeclareMathOperator*{\argmin}{arg\,min}

\usepackage{newfloat}
\usepackage{listings}
\DeclareCaptionStyle{ruled}{labelfont=normalfont,labelsep=colon,strut=off} 
\floatstyle{ruled}
\newfloat{listing}{tb}{lst}{}
\floatname{listing}{Listing}
\title{E2HQV: High-Quality Video Generation from Event Camera via Theory-Inspired Model-Aided Deep Learning}
\author{
    Qiang Qu\textsuperscript{\rm 1},
    Yiran Shen\textsuperscript{\rm 2}\thanks{Corresponding author. The implementation of our work is publicly available at https://github.com/VincentQQu/E2HQV.},
    Xiaoming Chen\textsuperscript{\rm 3}\footnotemark[1],
    Yuk Ying Chung\textsuperscript{\rm 1},
    Tongliang Liu\textsuperscript{\rm 1}
}
\affiliations{
    \textsuperscript{\rm 1}School of Computer Science, The University of Sydney.\\
    \textsuperscript{\rm 2}School of Software, Shandong University.\\ 
    \textsuperscript{\rm 3}School of Computer Science and Engineering, Beijing Technology and Business University.\\
    
    \{vincent.qu, vera.chung, tongliang.liu\}@sydney.edu.au; yiran.shen@sdu.edu; xiaoming.chen@btbu.edu.cn
}

\usepackage{bibentry}

\begin{document}

\maketitle

\begin{abstract}
The bio-inspired event cameras or dynamic vision sensors are capable of asynchronously capturing per-pixel brightness changes (called event-streams) in high temporal resolution and high dynamic range.
However, the non-structural spatial-temporal event-streams make it challenging for providing intuitive visualization with rich semantic information for human vision. It calls for  events-to-video (E2V) solutions which take event-streams as input and generate high quality video frames for intuitive visualization. However, current solutions are predominantly data-driven without considering the prior knowledge of the underlying statistics relating event-streams and video frames. It highly relies on the non-linearity and generalization capability of the deep neural networks, thus, is struggling on reconstructing detailed textures when the scenes are complex. 
In this work, we propose \textbf{E2HQV}, a novel E2V paradigm designed to produce high-quality video frames from events. This approach leverages a model-aided deep learning framework, underpinned by a theory-inspired E2V model, which is meticulously derived from the fundamental imaging principles of event cameras.
To deal with the issue of state-reset in the recurrent components of E2HQV, we also design a temporal shift embedding module to further improve the quality of the video frames. Comprehensive evaluations on the real world event camera datasets validate our approach, with E2HQV, notably outperforming state-of-the-art approaches, e.g., surpassing the second best by over 40\% for some evaluation metrics. 
\end{abstract}


\section{Introduction}

Inspired by the human visual system, Silicon Retina \cite{mahowald264mead} has pioneered an approach to perceptual sensing with event cameras or Dynamic Vision Sensors (DVS)~\cite{lichtsteiner2008128, posch2010qvga, berner2013240} and gained significant interests from both academia and industry.
Unlike traditional cameras, event cameras detect microsecond-level intensity changes, generating an asynchronous stream of `events', termed as event-stream. Event cameras offer several advantages over conventional CCD/CMOS cameras, including high temporal resolution, high dynamic range of up to 140dB~\cite{lichtsteiner2008128}, and low resource consumption due to the sparse nature of event-streams. For example, the DVS128 sensor platform consumes 150 times less energy than a conventional CMOS camera \cite{lichtsteiner2008128}.

Despite the appealing advantages of event cameras, the non-structural event-streams are not inherently compatible with traditional computer vision methodologies~\cite{scheerlinck2020fast} and the visualization is not intuitive for human users to understand. To address the above issue, the research on events-to-video (E2V), which aims to generate video frames from pure event-streams, has been raised to provide convenient and intuitive access to the rich information encapsulated in the sparse and non-structure event-streams. There have been a number of successful approaches for E2V task, such as  E2VID~\cite{rebecq2019high}, FireNet~\cite{scheerlinck2020fast}, SPADE-E2VID~\cite{cadena2021spade}, and ET-Net~\cite{weng2021event}.

\begin{figure}[!htb]
  \centering
   \includegraphics[width=0.9\linewidth]{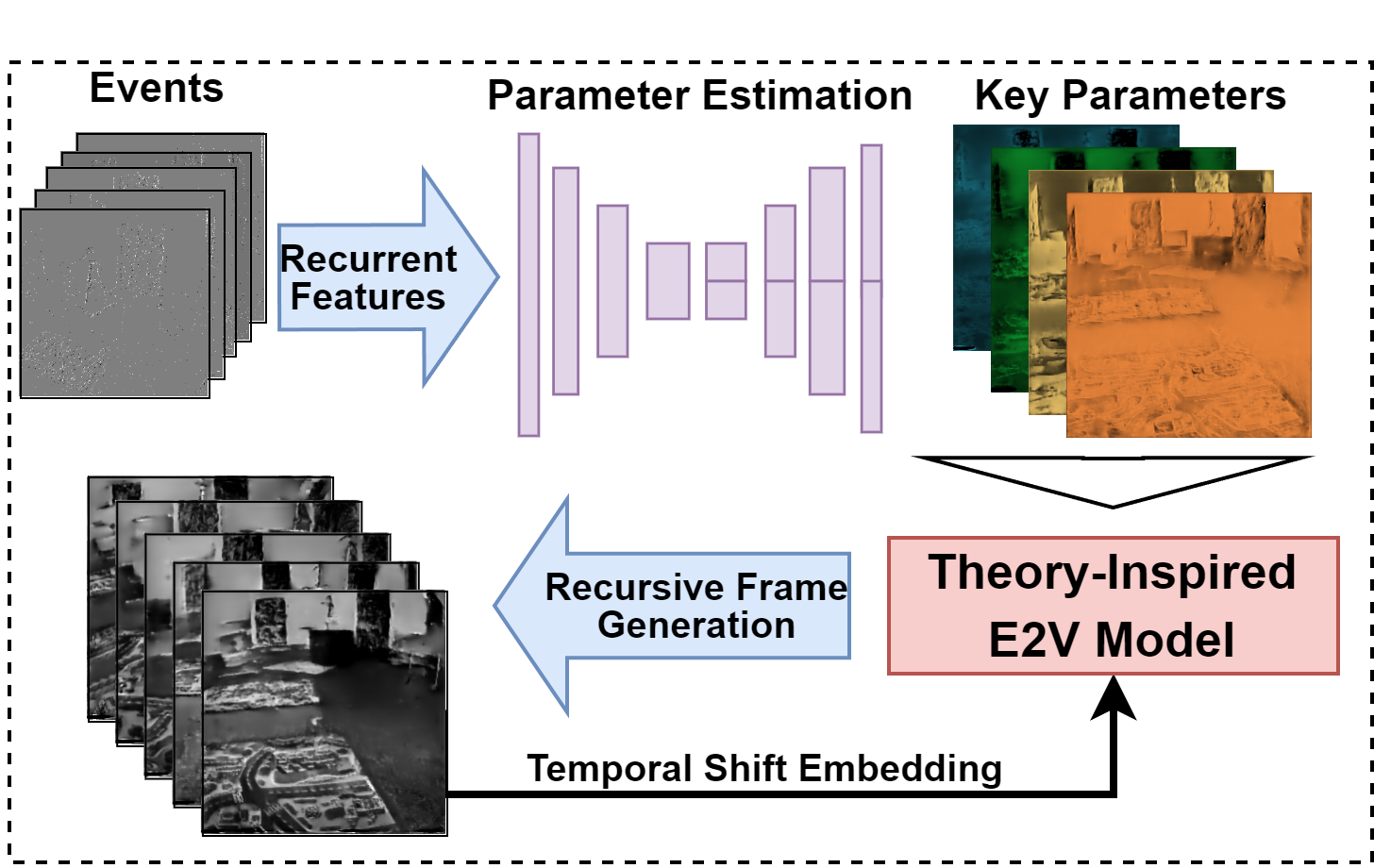}

   \caption{Conceptual Overview of the Proposed Model-Aided Learning Framework. }
   \label{fig:trigger}
\end{figure}

However, the quality of the video frames generated by the existing E2V approaches is still not satisfactory and fail to recover detailed texture for the complex scenes~\cite{ercan2023evreal}. This issue is predominantly attributed to the fact that many of these approaches, such as E2VID and ET-Net, primarily adopt a purely data-driven approach to learn the mapping from event-streams to video frames directly. However, the purely data-driven approaches are lack of interpretability and flexibility~\cite{shlezinger2023model}, and they do not take into account the prior knowledge of the underlying statistics relating event-streams and video frames. Therefore, their performance is largely dependent on the non-linearity and generalization capability of the neural networks, which poses significant challenges when the scenes to be reconstructed are complex~\cite{jarrett2009best}.

To address the aforementioned challenges, we introduce \textbf{E2HQV}, a novel E2V paradigm designed to produce high-quality video frames from events. This is achieved through a model-aided deep learning framework that integrates a theory-inspired E2V model. Rooted in the fundamental imaging principles of event cameras, this theory-inspired E2V model elucidates the relationship between consecutive frames and their associated inter-frame event-streams, offering valuable prior knowledge that enhances the learning efficacy of our deep learning framework. As shown in Figure~\ref{fig:trigger}, instead of generating video frames in a pure data-driven approach, E2HQV estimates a number of intermediate key parameters defined by the theory-inspired E2V model then reconstructs the video frames accordingly. The contributions of this work can be summarized as:

\begin{itemize}

\item We propose E2HQV, a novel high-quality video frames generation approach from event-streams by facilitating a model-aided learning framework which learns the key parameters defined by a theory-inspired E2V model and generates high quality video frames accordingly. 

\item According to the imaging principle of event camera and relation between video frames and event-stream, a theory-inspired E2V model is derived to guide the design of the model-aided learning framework.

\item A new temporal shift embedding module is designed to deal with the perturbation introduced by the state-reset mechanism of the recurrent components in the framework and ensuring seamless fusion of events and reconstructed frames.

\item Through extensive experiments on mainstream event-based video reconstruction datasets, E2HQV consistently exhibits superior performance over state-of-the-art (SOTA) approaches. Remarkably, for certain evaluation metrics, E2HQV surpasses the next best approach by a substantial margin of over 40\%.

\end{itemize}

\section{Related Works}

With the advent of deep learning, several methods have been proposed that use neural networks for event-based video reconstruction. Rebecq et al. \cite{rebecq2019high} proposed a ConvLSTM-based model that leverages the spatiotemporal representation of events for video reconstruction, providing high speed and dynamic range video with an event camera. FireNet \cite{scheerlinck2020fast} offers fast image reconstruction with a lightweight network. Stoffregen et al. \cite{stoffregen2020reducing} proposed an augmentation method on simulated training data that improves the performance of E2VID (E2VID+) and FireNet (FireNet+). SPADE-E2VID \cite{cadena2021spade} introduces spatially-adaptive denormalization for event-based video reconstruction. SSL-E2VID \cite{paredes2021back} focuses on self-supervised learning of image reconstruction via photometric constancy. Lastly, ET-Net \cite{weng2021event} utilizes a vision transformer for event-based video reconstruction, suffering from computational burden.

In our pursuit of a suitable deep learning framework to serve as the backbone for the proposed parametric frame generator, we conducted a review of several extant neural network-based architectures. Over the past decade, myriad architectures have been developed, including but not limited to ResNet \cite{szegedy2017inception}, MobileNet \cite{howard2017mobilenets}, SENet \cite{hu2018squeeze}, EfficientNet \cite{tan2019efficientnet}, and the more recently introduced EfficientNetV2 \cite{tan2021efficientnetv2}. These models have consistently set performance benchmarks, advancing the field substantially.
Among these models, we opted for EfficientNetV2 \cite{tan2021efficientnetv2}, which represents the SOTA in the field, due to its optimal balance between training time and parameter efficiency.
Elaborating on our choice, EfficientNetV2 \cite{tan2021efficientnetv2} has been crafted employing a blend of training-aware neural architecture search and scaling. This results in a model that demonstrates superior training speed and parameter efficiency compared to its predecessors. For the construction of our proposed parametric frame generator's backbone, we incorporated the MBConv layers and Fused MBConv layers from EfficientNetV2.

\section{Methodology}

\begin{figure}[!htb]
  \centering
   \includegraphics[width=0.85\linewidth]{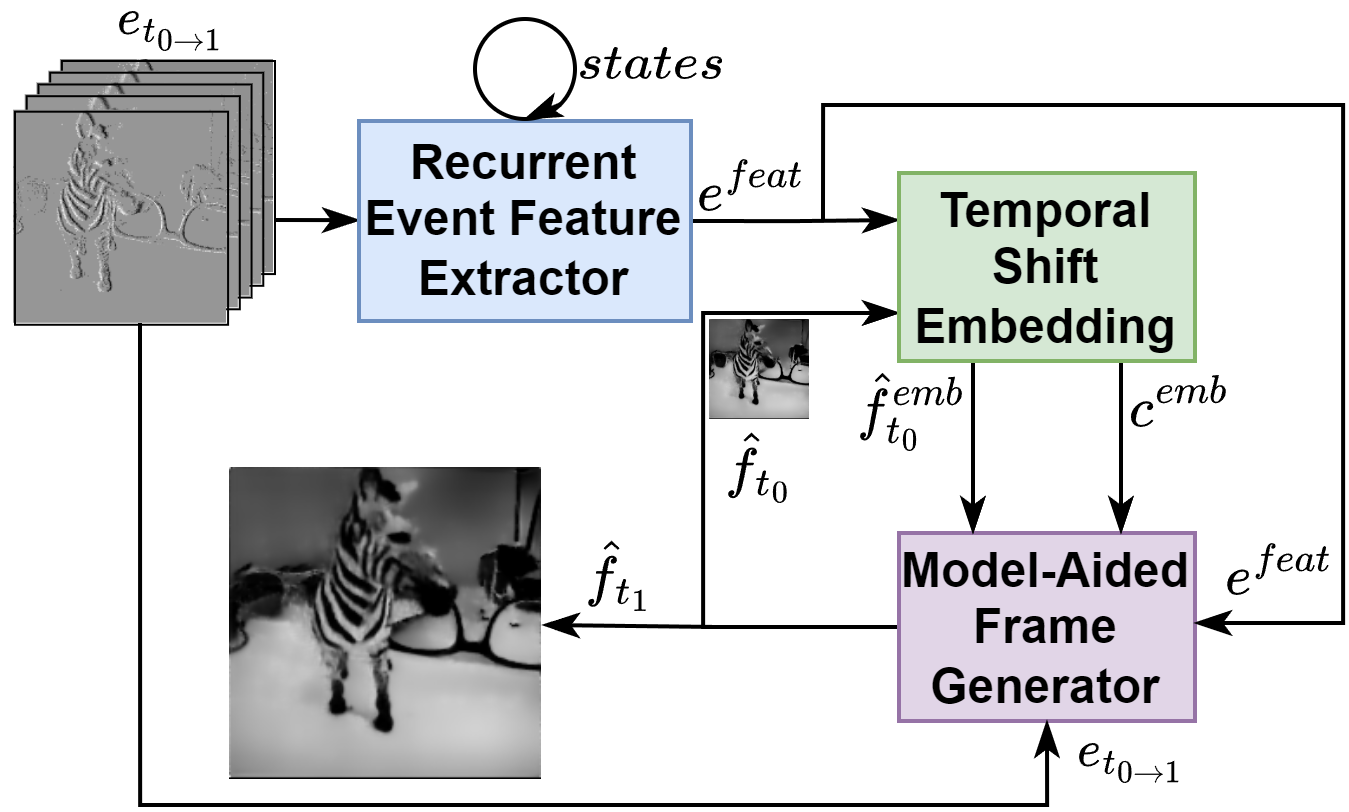}

   \caption{ Overview of the Proposed Model-Aided Learning Framework. }
   \label{fig:framework}
\end{figure}

\subsection{Overview of E2HQV}
As shown in Figure~\ref{fig:framework}, the pipeline of E2HQV can be vastly categorized into three major components, including a Recurrent Event Feature Extraction (REFE) module, a Temporal Shift Embedding (TSEM) module and a Model-Aided Frame Generation (MAFG) module. 
The {\bf REFE} module is implemented by leveraging a lightweight ConvLSTM UNet~\cite{shi2015convolutional} and it takes the event-stream $e_{t_{0\rightarrow1}^i}$, represented in VoxelGrid~\cite{zhu2019unsupervised} following SOTA E2V approaches~\cite{rebecq2019high, scheerlinck2020fast, weng2021event}, to extract the features effectively from the spatial-temporal event-stream. The {\bf TSEM} module is proposed to deal with the issue of state-reset in recurrent models. It takes the features extracted from the REFE module and the last frame generated from the MAFG module as input, and provide a new embedding for the following the MAFG module to deal with imprecise state-reset. The {\bf MAFG} module is the backbone of E2HQV. It exploits the theoretical relation between consecutive frames and event-stream in-between to determine the key parameters needed for E2V generation. Then a model-aid learning framework is proposed to estimate the key parameters and then generate video frames accordingly.

\noindent{\bf Problem Formulation.} Figure~\ref{fig:framework} also presents the inputs, outputs and intermediate results passed within E2HQV. In the context of E2V generation, a pair of consecutive generated frames can be expressed as $\hat{p}_i=\{ \hat{f}_{t_0^i}, \hat{f}_{t_1^i} \}$, corresponding to timestamp $t_0^i$ and $t_1^i$, respectively. The concurrent event-stream is denoted as $e_{t_{0\rightarrow1}^i}=\{e_t | t \in [t_0^i, t_1^i]\}$, are the set of events triggered during the time interval $[t_0^i,t_1^i]$. Provided the last generated frame $\hat{f}_{t_0^i}$ and the following event-stream $e_{t_{0\rightarrow1}^i}$, the objective of E2V generation is to learn an optimal model $H^*$ such that,

\begin{equation}
\label{eq:obj}
\begin{aligned}
H^* &= \argmin {\frac{1}{N}\sum_{i=1}^{N} loss[H\{e_{t_{0\rightarrow1}^i}, {\hat{f}_{t_0^i}}\},f_{t_1^i}]},
\end{aligned}
\end{equation}
where $N$ is the total number of frame pairs. Model $H$ takes event-stream $e_{t_{0\rightarrow1}^i}$ and the last reconstructed frame $\hat{f_{t_0^i}}$ as input to generate the next frame $\hat{f}_{t_1^i}$. The $loss$ function measures the difference between the reconstructed frame $\hat{f}_{t_1^i}$ and the ground truth $f_{t_1^i}$.

\begin{figure*}[htb]
  \centering
   \includegraphics[width=1.0\linewidth]{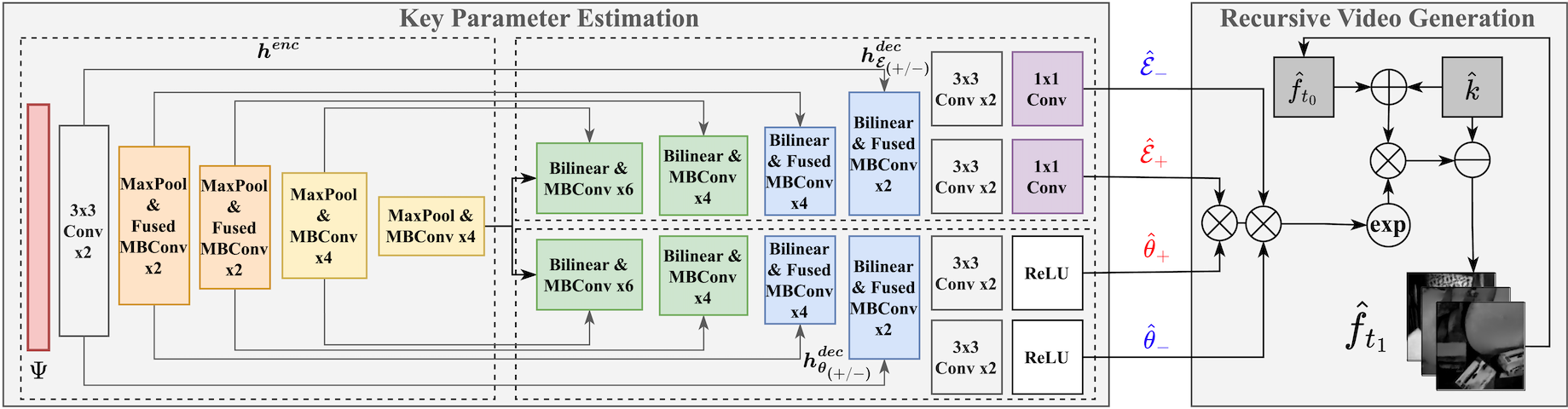}
   \caption{ The detailed settings of the model-aided frame generator (MAFG). 
   The generator accepts multimodal features $\Psi$ integrated from the output of the REFE module and the TSEM module, which are then input into a shared downsampling encoder, $h^{enc}$. This is followed by two upsampling decoders, $h^{dec}_{\mathcal{E}_{(+/-)}}$ and $h^{dec}_{\theta_{(+/-)}}$, and four output branches. These branches are meticulously designed for the estimation of the key parameters, $\mathcal{E_{(+/-)}}$ and $\theta_{(+/-)}$, respectively. Provided the estimated parameters, video frames are recursively generated from events according to Equation~\eqref{eq:f_n} derived from the theoretical relation between frame and event-stream.}
   \label{fig:eff_multiwnet}
\end{figure*}

As shown in Figure~\ref{fig:framework}, in each cycle of frame generation, the REFE module takes the event-stream $e_{t_{0\rightarrow1}}$ as input to extract recurrent event features, denoted as $e^{feat}$. Then the event features $e^{feat}$ and the last generated frame $\hat{f}_{t_0}$ are fed to TESM module to produce the embedded frame $\hat{f}_{t_0}^{emb}$ and a new embedding channel $c^{emb}$. The new channel indicates the relative distance to the last reset frame to aid the following MAFG module for generating high-quality video frames recursively with the integration of different features:
\begin{equation}
\label{eq:composite}
\begin{aligned}
\Psi = \{ \hat{f}_{t_0}^{emb}, c^{emb}, e_{t_{0\rightarrow1}}, e^{feat} \},
\end{aligned}
\end{equation}

\subsection{Model-Aided Frame Generation}
\label{subsec:frame_generation_with_parametric_estimation}
The backbone of our proposed E2HQV is a Model-Aided Frame Generator (MAFG). According to the theory-inspired E2V model derived from the theoretical relation between video frames and event-stream, it is designed as two major components: the key parameters estimation and recursive video generation.

\subsubsection{Theory-Inspired E2V Model}

Intuitively, event-streams signify brightness changes, while video frames directly record brightness levels, suggesting an intrinsic relationship between them. In this section, leveraging the imaging principles of event cameras, we derive a relationship between consecutive frames and their associated event-streams. This prior knowledge can then bolster the effectiveness of our deep learning framework, detailed in the next section.

According to the event generation model\cite{lichtsteiner2008128}, an event is triggered when a log-intensity change is detected (over a threshold). The log-intensity $d_i^{x,y}$ at pixel (x,y) can be expressed as,

\begin{equation}
  d_i^{x,y} = log I_{i+1}^{x,y} - log I_i^{x,y}.
  \label{eq:d_i}
\end{equation}
where $I_{i}^{x,y}$ and $I_{i+1}^{x,y}$ are the intensity values of pixel $(x,y)$ when two consecutive events are generated. The accumulated log-intensity change within $[t_0, t_1]$ (time interval between two consecutive video frames $f_{t_0}$ and $f_{t_1}$) can be approximated as,

\begin{equation}
\label{eq:sum_d_i}
\begin{aligned}
\sum_{i=0}^{n_{x,y}-1} d_i^{x,y} = log I_n^{x,y} - log I_{n-1}^{x,y} + log I_{n-1}^{x,y}\\ \quad ... - log I_1^{x,y} + log I_1^{x,y} - log I_0^{x,y}.
\end{aligned}
\end{equation}
By cancelling out the same items, the equation above can be rewritten as,
\begin{equation}
  \sum_{n=0}^{n_{x,y}-1} d_i^{x,y} = log I_n^{x,y} - log I_0^{x,y},
  \label{eq:sum_d_i_clean}
\end{equation}
Then $d_i^{x,y}$ can be decomposed as,

\begin{equation}
\label{eq:d_i_sub}
d_i^{x,y}=
    \begin{cases}
        + {{\theta}_{+}}_i^{x,y} & \text{if } d_i^{x,y} \geq 0 \\
        - {{\theta}_{-}}_i^{x,y} & \text{if } d_i^{x,y} < 0
    \end{cases},
\end{equation}
where ${{\theta_{(+/-)}}_i^{x,y}}$ is the polarity-wise absolute value of the threshold to trigger an event at pixel $(x,y)$.
As the period of $[t_0, t_1]$ is typically only tens of milliseconds, we assume $\theta_{(+/-)}^{x,y}$ is a constant threshold for pixel $p_{x,y}$ during the specific period.
The positive events (denoted as ``$+$'') and negative events (denoted as ``$-$'') are counted over $[t_0, t_1]$ at pixel $(x,y)$, i.e.,
\begin{equation}
\label{eq:ie}
\begin{aligned}
{\mathcal{E}_{(+/-)}}(x,y) = \sum_{i=0}^{n_{x,y}-1}{(+/-)}_i^{x,y}.
\end{aligned}
\end{equation}
Combing Eq.~\eqref{eq:d_i_sub} and Eq.~\eqref{eq:ie}, the accumulative log-intensity change within $[t_0, t_1]$ can be expressed as,
\begin{equation}
\label{eq:sum_d_i_c}
\begin{aligned}
\sum_{i=0}^{n_{x,y}-1} d_i^{x,y} &= {{\theta}_{+}}_i^{x,y} \sum_{i=0}^{n_{x,y}-1}{+}_i^{x,y}  - {{\theta}_{-}}_i^{x,y} \sum_{i=0}^{n_{x,y}-1}{-}_i^{x,y} \\ &= {\theta^{x,y}_{+}  {\mathcal{E}}^{x,y}_{+}-\theta^{x,y}_{-}  {\mathcal{E}}^{x,y}_{-}}.
\end{aligned}
\end{equation}
By incorporating Eq.~\eqref{eq:sum_d_i_clean} into Eq.~\eqref{eq:sum_d_i_c}, we have
\begin{equation}
  {\theta^{x,y}_{+}  {\mathcal{E}}^{x,y}_{+}-\theta^{x,y}_{-}  {\mathcal{E}}^{x,y}_{-}} = log \frac{I_n^{x,y}}{I_0^{x,y}}.
  \label{eq:c_ie}
\end{equation}
Then we assume a linear relationship between intensity $I$ and a corresponding normalized frame $f$, i.e.,

\begin{equation}
\label{eq:I_i}
\begin{aligned}
  I = af + b,
\end{aligned}
\end{equation}
where $a > 0$ and the pixel values of $f(x,y) \in [0, 1]$.
Then,
\begin{equation}
\label{eq:I_n_I_0}
\begin{aligned}
  \frac{I_n^{x,y}}{I_0^{x,y}} = \frac{af_{t_1}^{x,y} + b}{af_{t_0}^{x,y} + b} = \frac{f_{t_1}^{x,y} + k}{f_{t_0}^{x,y} + k},
\end{aligned}
\end{equation}
where $k = b/a$, $f_{t_0}$ and $f_{t_1}$ are normalized frames at timestamps $t_0$ and $t_1$.
Finally, based on Eq.~\eqref{eq:c_ie} and Eq.~\eqref{eq:I_n_I_0}, the APS frame $f_{t_1}$ can be represented as,
\begin{equation}
\label{eq:f_n}
\begin{aligned}
f_{t_1}^{x,y} = exp( {\theta^{x,y}_{+}  {\mathcal{E}}^{x,y}_{+}-\theta^{x,y}_{-}  {\mathcal{E}}^{x,y}_{-}} ) (f_{t_0}^{x,y} + k) - k.
\end{aligned}
\end{equation}

The equation above is our {\bf Theory-Inspired E2V Model} derived from imaging principal of event camera, and the relationship between frames and events. According to the model, the video frames can be generated recursively provided the key parameters \{$\theta_{(+/-)}$, k\} and the counts of positive and negative events $\mathcal{E}_{(+/-)}$.

However, because the thresholds of event camera are unknown and changing overtime, it is challenging to obtain the thresholds accurately. Then due to the noisy nature of event camera~\cite{wu2020probabilistic, baldwin2020event, guo2022low}, $\mathcal{E}_{(+/-)}$ obtained from simply counting the number of events overtime may introduce significant interference to the theory-inspired E2V model.

\subsubsection{Detailed Design of the MAFG}
To address the challenges above, we design a Model-Aided Frame Generator (MAFG) to estimate the key parameters and generate high-quality video frames from events according to the theory-inspired E2V model. As shown in Figure~\ref{fig:eff_multiwnet}, the MAFG facilities a deep neural network for key parameters estimation and then, provided estimated parameters and events, generates video frames recursively through simple addition and multiplication operations.

As shown in Fig.~\ref{fig:eff_multiwnet}, the backbone of the MAFG starts with taking the combination of features from other modules (refer to Figure~\ref{fig:framework}) as input to the encoder $h^{enc}$. The encoder consists of multiple convolutional and (Fused) MBConv~\cite{sandler2018mobilenetv2, tan2021efficientnetv2} layers followed by MaxPooling for feature extraction. The MBConv utilizes the inverted bottleneck structure \cite{sandler2018mobilenetv2} and depthwise convolutional layers \cite{howard2017mobilenets} to improve memory efficiency. In addition, a squeeze-and-excitation unit \cite{hu2018squeeze} is inserted in the middle of MBConv in order to adaptively recalibrate channel-wise feature responses. Then two non-sharing decoders $h^{dec}_{\mathcal{E}_{(+/-)}}$ and $h^{dec}_{\theta_{(+/-)}}$ with the same network architecture (except for the last two layers as shown in Fig.~\ref{fig:eff_multiwnet}) are adopted to decode the low-dimensional features with upsampling layers to obtain the estimates $\hat{\mathcal{E}}_{(+/-)}$ and $\hat{\theta}_{(+/-)}$. The decoder consists of the blocks of a combination of spatial bilinear interpolation and (Fused) MBConv layers. A $1\times 1$ convolutional layer without activation function is applied to finalize $h^{dec}_{\mathcal{E}_{(+/-)}}$ and ReLU activation function is adopted to ensure the non-negative property of $\theta_{(+/-)}$. At last, according to the theory-inspired E2V model (Eq.~\eqref{eq:f_n}), the estimate of the frame, $\hat{f}_{t_1}$, can be generated provided the estimated $\{\hat{\mathcal{E}}_{(+/-)}, \hat{\theta}_{(+/-)}\}$ and last generated frame $\hat{f}_{t_0}$.
The training loss is the Mean Absolute Error between the generated frames and the ground truths.

\begin{figure}[!htb]
  \centering
   \includegraphics[width=0.9\linewidth]{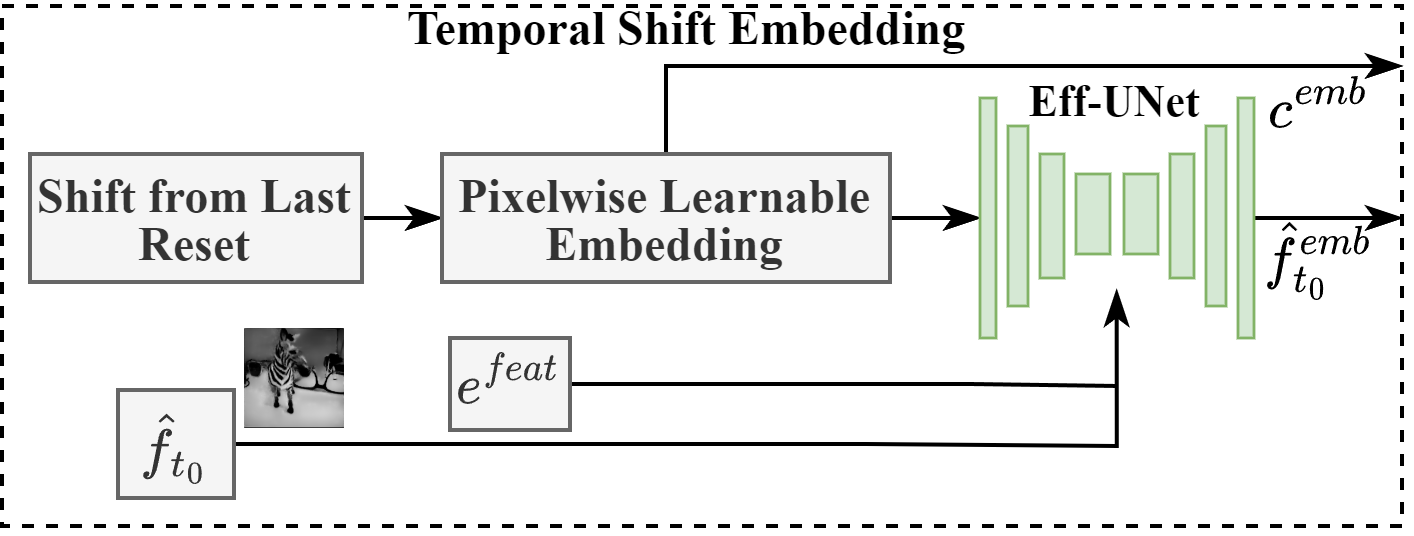}

   \caption{ Structure of the Temporal Shift Embedding.}
   \label{fig:temporal_embedding}
\end{figure}

\begin{table*}[htb]
\centering
\small
\renewcommand{\arraystretch}{1.0}
\setlength{\tabcolsep}{5pt}
\begin{tabular}{|l|ccc|ccc|ccc|}
\hline
& \multicolumn{3}{c|}{IJRR} & \multicolumn{3}{c|}{MVSEC } & \multicolumn{3}{c|}{HQF }  \\ \hline
& MSE↓ & SSIM↑ & LPIPS↓ & MSE↓ & SSIM↑ & LPIPS↓ & MSE↓ & SSIM↑ & LPIPS↓ \\ \hline
E2VID & 0.212 & 0.424 & 0.350 & 0.337 & 0.206 & 0.705 & 0.127 & 0.540 & 0.382  \\
FireNet& 0.131 & 0.502 & 0.320 & 0.292 & 0.261 & 0.700 & 0.094 & 0.533 & 0.441  \\
E2VID+ & 0.070 & 0.560 & 0.236 & 0.132 & 0.345 & 0.514 & 0.036 & 0.643 & \textbf{0.252}  \\
FireNet+ & 0.063 & 0.555 & 0.290 & 0.218 & 0.297 & 0.570 & 0.040 & 0.614 & 0.314  \\
SPADE-E2VID & 0.091 & 0.517 & 0.337 & 0.138 & 0.342 & 0.589 & 0.077 & 0.521 & 0.502  \\
SSL-E2VID  & \underline{0.046} & 0.364 & 0.425 & \underline{0.062} & 0.345 & 0.593 & 0.126 & 0.295 & 0.498 \\
ET-Net & 0.047 & \underline{0.617} & \underline{0.224} & 0.107 & \underline{0.380} & \underline{0.489} & \underline{0.032} & \underline{0.658} & \underline{0.260} \\ \hline
E2HQV (Ours) & \textbf{0.028} & \textbf{0.682} & \textbf{0.196} & \textbf{0.032} & \textbf{0.421} & \textbf{0.460} & \textbf{0.019} & \textbf{0.671} & 0.261 \\ \hline
\end{tabular}
\caption{ Quantitative comparisons of the evaluated SOTA methods on IJRR, MVSEC, and HQF. The best results are in bold while the second best results are underlined (the same for the rest tables).}
\label{tab:benchmarking}
\end{table*}

\subsubsection{Temporal Shift Embedding Module}
\label{subsec:temporal_shift_embedding}

The theory-inspired E2V model, as depicted in Eq.~\eqref{eq:f_n}, recursively takes previously synthesized frames and concurrent events to generate succeeding frames. Periodic state-resets, commencing at the outset and recurring after each predefined frame generation interval, are imperative due to issues like interference between unrelated sequences and the inconsistency of the starting state~\cite{chung2014empirical, le2015simple}. This could potentially impact the performance of the proposed method. The MAFG module, devoid of knowledge pertaining to its relative temporal distance from the last reset frame, may lack the capability for adaptive fusion of events with the most recent reconstructed frame.

As shown in overview of the E2HQV framework (Figure~\ref{fig:framework}), to resolve issue of state-reset, we introduce a Temporal Shift Embedding (TSEM) module designed to embed the previous frame and generate a new embedded channel to provide state-reset information for the MFGA module. This allows MFGA module to adjust its reconstruction behavior based on the position of current timestamp within the reset interval. Fig. \ref{fig:temporal_embedding} presents the detailed structure of the TSEM module. This module maintains a sequence of pixelwise learnable embedding tensors, each of which corresponds to a specific shift from the last state-reset within a reset interval. When provided with the shift from the last reset, the TSEM module identifies the relevant pixelwise embedding.
This embedding is then integrated with the recurrent event feature $e^{feat}$ and the previously generated frame $\hat{f}_{t_0}$, and fed into a lightweight and efficient UNet~\cite{shi2015convolutional}, namely Eff-UNet, to generate the embedded frame $\hat{f}_{t_0}^{emb}$ and produce a new embedding channel $c^{emb}$.
The Eff-UNet features an encoder-decoder structure reminiscent of $h^{enc}$ and $\mathcal{E_{+}}$ with significantly reduced convolutional filters at each layer, rendering it highly efficient.

\section{Experiments}

\subsection{Training on Simulated Dataset}
We utilize the simulated event-based dataset~\cite{rebecq2019high} converted from MS-COCO~\cite{cocodataset} for training which  is pervasively adopted in a number of E2V approaches~\cite{rebecq2019high, scheerlinck2020fast, cadena2021spade, weng2021event}. The dataset is generated using the ESIM event simulator \cite{mueggler2017event}, which produces synthetic events by rendering images from the MS-COCO dataset along a simulated camera trajectory. The camera sensor size is set to be 240 × 180 pixels, matching the resolution of the DAVIS240C sensor~\cite{berner2013240}.
Data enrichment is achieved by assigning different positive and negative contrast thresholds to each simulated scene, sampled from a normal distribution. This prevents the network from merely integrating events and ensures better generalization to real event-streams.
The dataset comprises 1,000 sequences of 2-seconds event-streams, totaling approximately 35 minutes.
For all the experiments, we utilize a desktop computer with an AMD 5950X processor, RTX 3090 GPU, and 32GB RAM, running on Ubuntu 20.04 and implemented using the PyTorch~\cite{NEURIPS2019_9015}.

\begin{figure*}[!htb]
  \centering

   \includegraphics[width=0.92\linewidth]{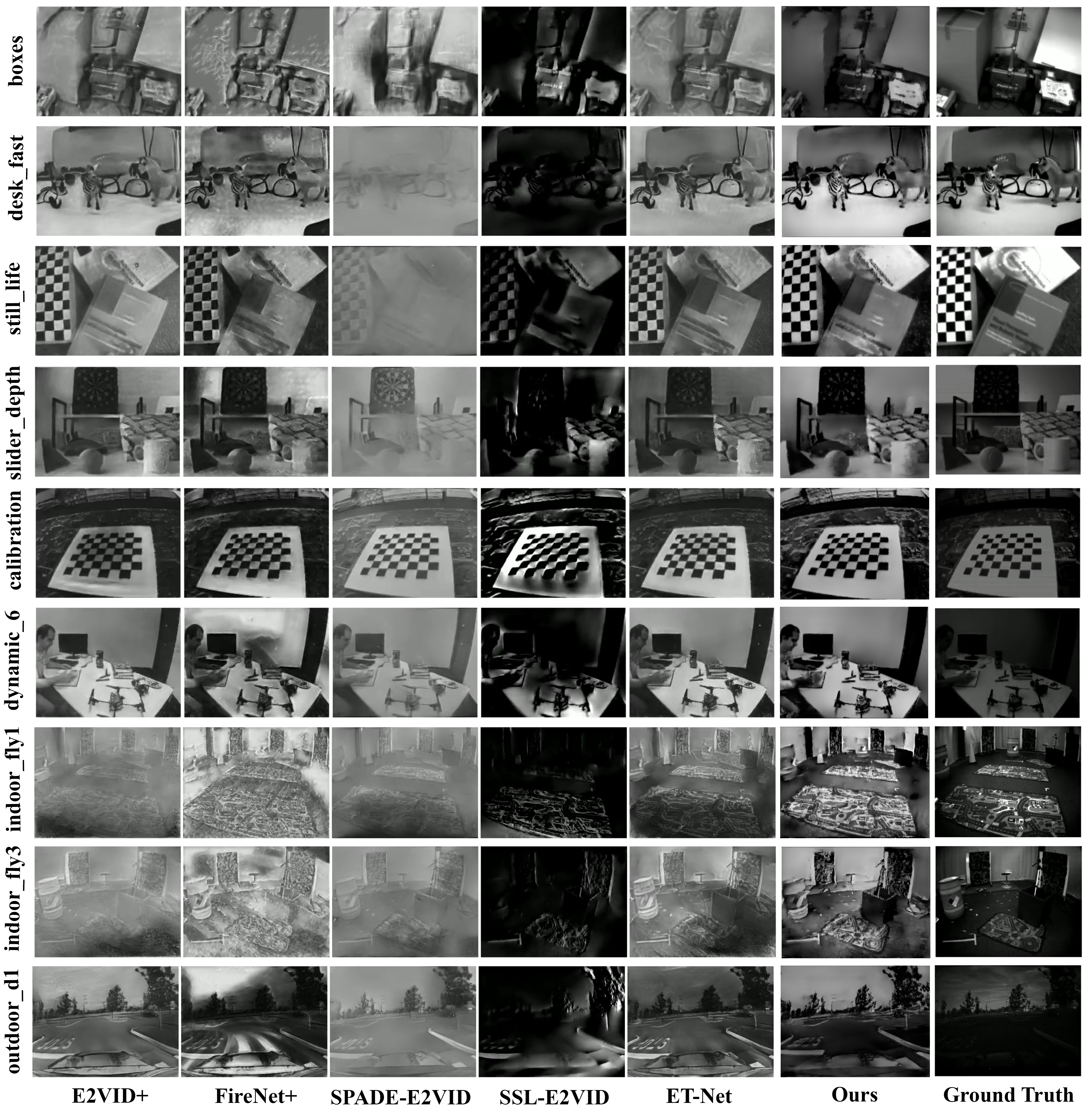}
   \caption{ Qualitative Analysis across Datasets. Comparative visualizations of sequence data from HQF (rows 1-3), IJRR (rows 4-6), and MVSEC (rows 7-9). The evaluated baseline methods often exhibit limitations such as diminished contrast, noticeable blur, and prominent artifacts. In contrast, our reconstructions offer high contrast and are adept at maintaining sharp edge details, while manifesting minimal artifacts in regions devoid of texture.
 }
   \label{fig:demo}
\end{figure*}

\subsection{Evaluation on Real World Datasets}
Our approach E2HQV is evaluated on three publicly available datasets consisting of real event-streams collected by event cameras, i.e., IJRR \cite{rebecq2019high}, MVSEC \cite{zhu2018multivehicle}, and HQF \cite{stoffregen2020reducing} which are commonly used to benchmark the accuracy of E2V approaches~\cite{rebecq2019high, scheerlinck2020fast, stoffregen2020reducing, paredes2021back, weng2021event}. 
To ensure strict consistency between the timestamps of the reconstruction and ground truth, we utilize the events between two consecutive frames to generate the later frame. Following the most comprehensive work on E2V benchmarking~\cite{ercan2023evreal}, E2HQV is compared with seven SOTA methods: E2VID~\cite{rebecq2019high}, FireNet~\cite{scheerlinck2020fast}, E2VID+~\cite{stoffregen2020reducing}, FireNet+~\cite{stoffregen2020reducing}, SPADE-E2VID~\cite{cadena2021spade}, SSL-E2VID~\cite{paredes2021back}, and ET-Net~\cite{weng2021event}. All the approaches are solely trained with simulated dataset and tested directly on the three real-world datasets without any further fine-tuning.
The accuracy of the generated frames are evaluated by comparing with the ground truths using the following metrics: Mean Squared Error (MSE), Structural Similarity (SSIM) \cite{wang2004image}, and Perceptual Similarity (LPIPS)~\cite{zhang2018unreasonable} which are the same as those in the literature.
\subsubsection{Quantitative Evaluation}
The quantitative evaluation results are presented in Table~\ref{tab:benchmarking}. From the results we can find, E2HQV demonstrates superior performance over the seven SOTA methods across all three metrics on the IJRR and MVSEC datasets. It also surpasses all other methods in terms of MSE and SSIM. Specifically, E2HQV reduces the MSE by approximately $40.4\%$, $70.1\%$, and $40.6\%$ across all the three datasets, compared with the second best approach, SSL-E2VID. In terms of SSIM, E2HQV significantly outperforms the second best approach, ET-Net, achieving scores of 0.682, 0.421, and 0.671 on the HQF, IJRR, and MVSEC datasets, respectively. For the Learned LPIPS metric, E2HQV either exceeds or is close to the best performance of SOTA methods.

\begin{table*}[htb]
\centering
\small
\begin{tabular}{|c|c|ccc|ccc|ccc|}
\hline
\multicolumn{2}{|c|}{Reset Interval} & \multicolumn{3}{c|}{IJRR} & \multicolumn{3}{c|}{MVSEC} & \multicolumn{3}{c|}{HQF} \\ \hline
REFE & TSEM & MSE↓ & SSIM↑ & LPIPS↓ & MSE↓ & SSIM↑ & LPIPS↓ & MSE↓ & SSIM↑ & LPIPS↓ \\ \hline
20 & 20 & 0.032 & \underline{0.665} & \underline{0.193} & \underline{0.040} & \textbf{0.447} & \textbf{0.442} & \underline{0.020} & \underline{0.670} & 0.273 \\
20 & 40 & \underline{0.029} & 0.656 & \textbf{0.190} & 0.049 & 0.419 & 0.463 & 0.023 & 0.663 & 0.277 \\
40 & 20 & \textbf{0.028} & \textbf{0.682} & 0.196 & \textbf{0.032} & \underline{0.421} & \underline{0.460} & \textbf{0.019} & \textbf{0.671} & \textbf{0.261} \\
40 & 40 & 0.041 & 0.648 & 0.205 & 0.043 & 0.416 & 0.475 & 0.021 & 0.667 & \underline{0.271} \\ \hline
\end{tabular}
\caption{Ablation study on different combination of reset intervals.}
\label{tab:memory_length}
\end{table*}

\subsubsection{Qualitative Evaluation}

Figure~\ref{fig:demo} provides qualitative comparison of the generated video frames from E2HQV and other competing methods across the three evaluated datasets. The ground truth is presented in the rightmost column as reference. By comparing the examples from different approaches, we can observe E2HQV shows superior performance on reconstructing complex scenes: it effectively reduces foggy artifacts (as evident in the first, second, third, fourth, and penultimate rows), enhances the details in the context (as seen in the tree details of the last row), and improves contrast (as observed in the third, fifth, penultimate, and last rows).

\begin{table*}[htb]
\centering
\small
\begin{tabular}{|l|ccc|ccc|ccc|}
\hline
& \multicolumn{3}{c|}{IJRR} & \multicolumn{3}{c|}{MVSEC} & \multicolumn{3}{c|}{HQF} \\ \hline
Selected Modules& MSE↓ & SSIM↑ & LPIPS↓ & MSE↓ & SSIM↑ & LPIPS↓ & MSE↓ & SSIM↑ & LPIPS↓ \\ \hline
MAFG & 0.080 & 0.575 & 0.263 & \underline{0.062} & 0.384 & \underline{0.465} & 0.033 & 0.631 & 0.310 \\
MAFG+REFE & 0.051 & 0.639 & 0.218 & 0.064 & 0.383 & 0.500 & 0.027 & \underline{0.653} & \underline{0.290} \\
MAFG+TSEM & \underline{0.032} & \underline{0.652} & \underline{0.202} & 0.072 & \underline{0.400} & 0.478 & \underline{0.025} & 0.639 & 0.309 \\
MAFG+REFE+TSEM & \textbf{0.028} & \textbf{0.682} & \textbf{0.196} & \textbf{0.032} & \textbf{0.421} & \textbf{0.460} & \textbf{0.019} & \textbf{0.671} & \textbf{0.261} \\ \hline
\end{tabular}
\caption{Ablation study on different combination of modules.}
\label{tab:different_module}
\end{table*}

\subsection{Ablation Study}

\subsubsection{Investigating State-Reset Intervals for REFE and TSEM}
The state-reset interval, as introduced in Section \textbf{Temporal Shift Embedding}, serves as a pivotal hyperparameter in TSEM. Furthermore, the REFE, which incorporates ConvLSTM units, also relies on the state-reset interval as a hyperparameter. This is due to the fact that these recurrent units are initialized to zero at each state-reset interval.
Given the presence of these two hyperparameters, it is of significant interest to conduct an ablation study to explore the impact of different combinations of these parameters to find the optimal configuration. 
The results, presented in Table \ref{tab:memory_length}, 
show that a combination of 40 and 20 for the REFE and the TSEM modules respectively achieves the best overall accuracy.

\subsubsection{Contributions of Different Modules}
To evaluate the effectiveness of each module of E2HQV, we conduct ablation studies on different combinations of the three modules. The results are reported in Table \ref{tab:different_module}. By comparing different configurations, the highest performance is achieved when all three modules are deployed and each module contributes to the video frames generation with a good margin. Especially, the comparison between the ``MAFG''  and ``MAFG+TSEM'' configurations reveals significant performance uplift, underscoring the substantial contributions of the TSEM module to the efficacy of E2HQV.

\begin{figure}[!htb]
  \centering
   \includegraphics[width=1\linewidth]{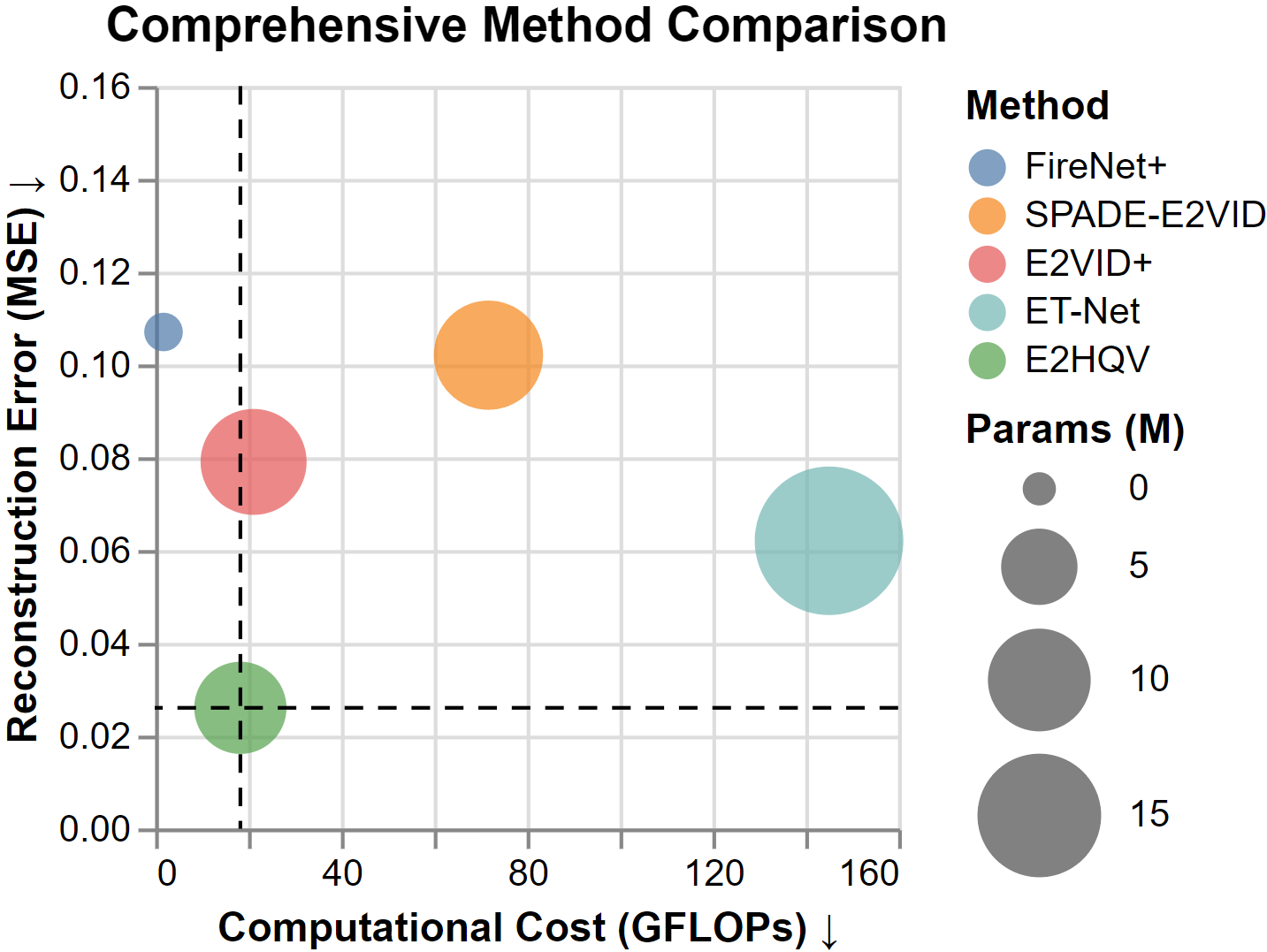}
   \caption{ Trade-off Between Accuracy and Complexity. 
   The reconstruction error (MSE) is shown on the y-axis 
   and the computational cost (GFLOPs) is depicted on the x-axis. 
   The number of parameters for each model is visually represented by the size of the corresponding circle.}
   \label{fig:trade_off}
\end{figure}

\begin{table}[htb]
\centering
\small
\renewcommand{\arraystretch}{1.}
\setlength{\tabcolsep}{5pt}
\begin{tabular}{|l|c|c|}
\hline
& Params (M) ↓ & GFLOPs ↓ \\ \hline
E2VID & 10.71 & 21.2 \\
FireNet & \textbf{0.04} & \textbf{1.8} \\
SPADE-E2VID & 11.46 & 71.78 \\
ET-Net & 22.18 & 145.12 \\
E2HQV (Ours) & \underline{7.82} & \underline{18.35} \\ \hline
\end{tabular}
\caption{Quantitative comparisons of model complexity}
\label{tab:complexity}
\end{table}

\subsubsection{Computational Complexity Analysis}
We also analyse the computational complexity of E2HQV, compared with the competing methods. To quantify the complexity of the approaches, we consider two salient computational metrics delineated in Table \ref{tab:complexity}, which are the number of model parameters and GFLOPs. Despite FireNet exhibiting the lowest model complexity, our proposed model attains a superior trade-off, balancing commendable accuracy against a judicious use of parameters and computational resources (visualized in Fig.~\ref{fig:trade_off}).

\section{Conclusion}
In this study, we propose E2HQV to improve the quality of generated video frames significantly for E2V task via model-aided learning with a theory-inspired E2V model derived from the imaging principle of event cameras. E2HQV, which integrates theoretical insights with data-driven learning, has shown superior performance over SOTA E2V approaches in extensive evaluations on real world datasets. The introduction of the temporal shift embedding module further enhances the robustness of our approach, ensuring seamless event and frame fusion. The comparison of results of different approaches in evaluation shows E2HQV can generate high quality video frames.

\bibliography{aaai24}

\end{document}